\documentclass[letterpaper, 10 pt, conference]{ieeeconf}  % Comment this line out if you need a4paper

\IEEEoverridecommandlockouts                              % This command is only needed if 
\usepackage{amsmath} % assumes amsmath package installed  数学公式支持
\usepackage{xcolor}
\usepackage{amssymb}  % assumes amsmath package installed  数学符号支持
\usepackage{soul}
\usepackage{graphicx}
\usepackage{subcaption} % 用于子图标注
\usepackage{caption}
\usepackage[font=small,labelfont=bf]{caption}
\usepackage{hyperref}
\usepackage{multirow}
\usepackage{booktabs}
\usepackage{makecell} % 用于单元格换行
\usepackage{algorithm} % 算法环境
\usepackage{algorithmic} % 算法内容
\usepackage{xr}

\title{\LARGE \bf
Semantic-LiDAR-Inertial-Wheel Odometry Fusion for Robust Localization in Large-Scale Dynamic Environments
}

\author{Haoxuan Jiang, Peicong Qian, Yusen Xie, Linwei Zheng, Xiaocong Li,\\ Ming Liu, and Jun Ma, \textit{Senior Member, IEEE}% <-this % stops a space
\thanks{Haoxuan Jiang, Yusen Xie, and Linwei Zheng are with the Robotics and Autonomous Systems Thrust, The Hong Kong University of Science and Technology (Guangzhou), 
        Guangzhou 511453, China (e-mail: hjiangax@connect.hkust-gz.edu.cn; yxie827@connect.hk-gz.edu.cn; lzhengad@connect.ust.hk).}%
\thanks{Peicong Qian and Ming Liu are with Shenzhen Unity Drive Innovation Technology Co., Ltd.,
        Shenzhen 518063, China (e-mail: epsilonjohn9527@gmail.com; liu.ming.prc@gmail.com).}%
\thanks{Xiaocong Li is with the College of Information Science and Technology, Eastern
Institute of Technology, Ningbo, Ningbo 315200, China (e-mail: xiaocongli@eitech.edu.cn). }
\thanks{Jun Ma is with the Robotics and Autonomous Systems Thrust, The Hong Kong University of Science and Technology (Guangzhou), Guangzhou 511453, China, and also with the Cheng Kar-Shun Robotics Institute, The Hong Kong University of Science and Technology, Hong Kong SAR, China (e-mail: jun.ma@ust.hk).} 
}
\begin{document}

\maketitle

\thispagestyle{empty}
\pagestyle{empty}

%%%%%%%%%%%%%%%%%%%%%%%%%%%%%%%%%%%%%%%%%%%%%%%%%%%%%%%%%%%%%%%%%%%%%%%%%%%%%%%%
\begin{abstract}

% Reliable low-drift global localization is crucial for large-scale, unbounded and dynamic outdoor applications. 
Reliable, drift-free global localization presents significant challenges yet remains crucial for autonomous navigation in large-scale dynamic environments. 
In this paper, we introduce a tightly-coupled Semantic-LiDAR-Inertial-Wheel Odometry fusion framework, which is specifically designed to provide high-precision state estimation and robust localization in large-scale dynamic environments. 
Our framework leverages an efficient semantic-voxel map representation and employs an improved scan matching algorithm, which utilizes global semantic information to significantly reduce long-term trajectory drift. 
Furthermore, it seamlessly fuses data from LiDAR, IMU, and wheel odometry using a tightly-coupled multi-sensor fusion Iterative Error-State Kalman Filter (iESKF). This ensures reliable localization without experiencing abnormal drift. 
Moreover, to tackle the challenges posed by terrain variations and dynamic movements, we introduce a 3D adaptive scaling strategy that allows for flexible adjustments to wheel odometry measurement weights, thereby enhancing localization precision. 
This study presents extensive real-world experiments conducted in a one-million-square-meter automated port, encompassing 3,575 hours of operational data from 35 Intelligent Guided Vehicles (IGVs). The results consistently demonstrate that our system outperforms state-of-the-art LiDAR-based localization methods in large-scale dynamic environments, highlighting the framework's reliability and practical value.

\end{abstract}
\section{INTRODUCTION}

Global localization in large-scale dynamic environments remains a formidable challenge, particularly in settings like ports and industrial parks. Traditional localization systems often depend on signal-based methods like the Global Positioning System (GPS) or WiFi. However, these approaches can be unreliable in scenarios where stable signal delivery is disrupted, such as during signal loss or transmission interruptions.
In recent years, Simultaneous Localization and Mapping (SLAM)~\cite{5940562,
% 10428583,
10287946} has witnessed rapid advancements in high-precision localization, which becomes indispensable for industrial applications of robots and autonomous systems. 

Sensor fusion-based SLAM methods typically achieve localization by aligning scanned frames with pre-constructed maps, but they often suffer from abnormal localization drift.
To mitigate this, some advanced approaches incorporate continuous odometry estimation within localization frameworks. These methods integrate IMU data, LiDAR scans, and wheel odometry into a joint localization framework, yielding improved results. However, the absence of semantic information still makes them vulnerable to long-term drift.
In conventional map representation, they typically rely on explicit feature representations like point clouds~\cite{dellenbach2022cticprealtimeelasticlidar, 10015694}, surfels~\cite{8967704}
% {Behley2018EfficientSS, 8967704, wang2019real}
, and voxels~\cite{9560947, liu2024voxel}. While effective in static environments, these methods struggle with sparse observations, occlusions, and dynamic scenes, frequently failing to distinguish dynamic objects from static backgrounds. 
Moreover, they lack semantic information, leading to challenges such as long-term drift and difficulties in re-localization upon system restart.
Additionally, traditional wheel odometry~\cite{9765591, 10342258}
% , 10026859}
encounters difficulties in complex terrains, such as slopes or slippery surfaces, where tire slippage or poor ground contact results in inaccurate motion estimation.

To address aforementioned issues, this paper proposes a tightly-coupled semantic-LiDAR-inertial-wheel odometry fusion framework designed to significantly enhance localization accuracy and robustness in dynamic, large-scale environments. The framework incorporates LiDAR, IMU, and wheel odometry data within a tightly coupled multi-sensor fusion scheme based on the iterative error state Kalman filter (iESKF)~\cite{9372856, 9697912}, which effectively mitigates motion distortion and abnormal localization drift.
Our framework employs a voxel-based semantic matching algorithm that tightly integrates semantic information with spatial geometric features to enhance environmental understanding and reduce long-term trajectory drift. By extracting semantic labels from LiDAR point clouds and mapping them onto a voxelized map, the algorithm effectively distinguishes dynamic objects from static environmental features, minimizing the impact of dynamic objects on localization results while maximizing the contribution of static objects to improve localization accuracy and stability.
% Our proposed iSVM facilitates rapid neighborhood queries and efficient storage, while the integration of semantic information enhances adaptability to dynamic environments, reduces noise, and improves environmental understanding. 
Furthermore, a 3D adaptive scaling strategy is proposed to address errors due to complex terrains. This strategy dynamically adjusts the weight of wheel speed observations based on motion states and terrain features, optimizing wheel odometry performance and ensuring improved adaptability across diverse and challenging terrains.

In summary, the main contributions of this paper are as follows:
\begin{itemize}

\item We propose a tightly-coupled semantic-LiDAR-inertial-wheel odometry fusion framework based on iESKF. This framework effectively integrates LiDAR, IMU, and wheel odometry data to mitigate motion distortion and abnormal localization drift.
% , significantly improving localization accuracy and robustness in dynamic, large-scale environments.

\item We introduce a semantic voxel-based matching algorithm that integrates LiDAR semantic labels with spatial geometric features to distinguish diverse dynamic objects from static environments, effectively reducing the long-term trajectory drift. 
% By minimizing the impact of dynamic objects and leveraging static features, the algorithm effectively reduces long-term trajectory drift.

\item We present a 3D adaptive scaling strategy to address errors caused by complex terrains. This strategy dynamically adjusts the weight of wheel speed observations based on motion states and terrain features, optimizing wheel odometry performance across diverse and challenging environments.

\item Our algorithm has been successfully deployed in a one-million-square-meter automated port, delivering precise and stable localization for 35 Intelligent Guided Vehicles (IGVs), demonstrating its reliability and effectiveness in complex real-world applications.

\end{itemize}

\section{RELATED WORK}
\subsection{Efficient Map Representation}

Map representation and scan registration serve as the fundamental basis for ensuring reliable localization in a LiDAR-based odometry or SLAM system. 
Explicit mapping directly utilize geometric data collected by sensors for map construction. Point cloud representations (e.g., CT-ICP~\cite{dellenbach2022cticprealtimeelasticlidar}, KISS-ICP~\cite{10015694}) directly utilize sensor data, offering high precision and intuitive geometric characteristics. However, they require significant storage and have low query efficiency, limiting their deployment in large-scale or real-time applications. Surfel representations (e.g., 
% SuMa~\cite{Behley2018EfficientSS}, 
SuMa++~\cite{8967704}) store local surface information, such as points and normals, reducing storage requirements and improving query efficiency while retaining some geometric details. However, they struggle with adaptability in sparse or dynamic environments, often losing fine details. Voxel-based representations (e.g., LiTAMIN2~\cite{9560947}, Voxel-SLAM~\cite{liu2024voxel}) leverage spatial partitioning for efficient querying and updating, making them suitable for dynamic and large-scale environments. However, their accuracy is determined by voxel resolution: low resolutions result in a loss of detail, while high resolutions greatly increase computational costs.
% \hl{However, their accuracy depends on voxel resolution-low resolutions lose detail, while high resolutions increase computational costs.}

To enhance the efficiency of processing and storing explicit map representations, various advanced data structures have been developed.
Incremental KD-Tree (iKD-Tree)~\cite{9697912, cai2021ikd} enables fast point cloud updates and efficient nearest neighbor queries in dynamic environments.
% iKD-Tree is a high-performance spatial indexing method for dynamic scenes, capable of rapidly adjusting its structure during point cloud updates. Compared to traditional KD-Tree~\cite{9372856, 10.1145/361002.361007}, iKD-Tree offers superior performance in point insertion and deletion, making it suitable for dynamic environments and efficient nearest neighbor queries. 
Incremental Voxel Map (iVox) from Faster-LIO~\cite{9718203} updates voxel grids incrementally, integrating probabilistic and geometric data for accurate, robust mapping and efficient storage management.
% iVox further optimizes voxel grid updates through incremental construction and real-time adjustment, integrating probabilistic and geometric information to improve map accuracy and robustness, while enabling efficient storage management. 
Voxelized Generalized Iterative Closest Point (GICP)~\cite{9560835} and Voxel-based Surface Covariance Estimator (VSCE) from iG-LIO~\cite{10380742} leverage voxelized point distributions to robustly estimate surface covariances, avoiding costly nearest neighbor searches. 
% VGICP combines the strengths of GICP~\cite{Segal2009GeneralizedICP} and Normal Distributions Transform (NDT)~\cite{1249285}, 
Voxelized GICP enhances the capabilities of GICP~\cite{Segal2009GeneralizedICP} through voxelization, 
maintaining reliability in sparse data scenarios and supporting efficient parallel optimization. Similarly, VSCE improves efficiency in processing dense scans while maintaining high accuracy in sparse and small field-of-view scenarios, and also enabling parallel optimization. However, these methods primarily focus on geometric information and struggle to effectively distinguish between static and dynamic objects, often leading to localization drift in dynamic environments.

\subsection{Multi-Sensor Fusion based Localization}

Based on reliable and efficient mapping, LiDAR-based odometry and SLAM methods utilize sensors like LiDAR, IMU, and wheel odometry to achieve highly precise localization. 
Within the LiDAR-inertial odometry (LIO) framework, FAST-LIO~\cite{9372856} employs a tightly-coupled Kalman filter to reduce computational overhead and correct motion distortion, enabling robust navigation in dynamic environments. 
Building on this, FAST-LIO2~\cite{9697912} improves accuracy through raw point-to-map registration and boosts efficiency by utilizing iKD-Tree~\cite{cai2021ikd} for efficient map management and querying. 
Faster-LIO~\cite{9718203} further improves performance by replacing iKD-Tree~\cite{cai2021ikd} with iVox for faster updates and approximate k-NN queries, avoiding complex tree operations. 
Point-LIO~\cite{Point-LIO} adopts a point-by-point framework for high-frequency odometry updates, effectively removing motion distortion, and introduces a stochastic process-augmented kinematic model for accurate localization during aggressive motions, even in scenarios with IMU saturation. 
Beyond standard LIO frameworks, methods like EKF-LOAM~\cite{9765591}, LIWO~\cite{10342258}, and LIWOM-GD~\cite{10609751} integrate additional wheel odometry data to further enhance state estimation. 
EKF-LOAM enhances LeGO-LOAM~\cite{8594299} by employing an adaptive Extended Kalman Filter (EKF) with a lightweight covariance scheme to improve path estimation in feature-sparse environments, while LIWO utilizes a tightly-coupled bundle adjustment (BA) framework to achieve more accurate velocity estimation and effectively mitigate IMU drift.
LIWOM-GD~\cite{10609751} further incorporates an advanced dynamic point removal technique, global plane constraints and loop closure, all within a refined factor graph optimization framework. These works show satisfactory performance in static environments. However, their effectiveness declines considerably in dynamic and repetitive settings, particularly when encountering sparse observations and unforeseen collisions.
%They excel in static environments but face challenges in sparse observations, occlusions, and repetitive scenes, causing localization noise accumulation and instability.
\section{METHODOLOGY}

\begin{figure*}[htbp]
    \centering
    \includegraphics[width=\linewidth]{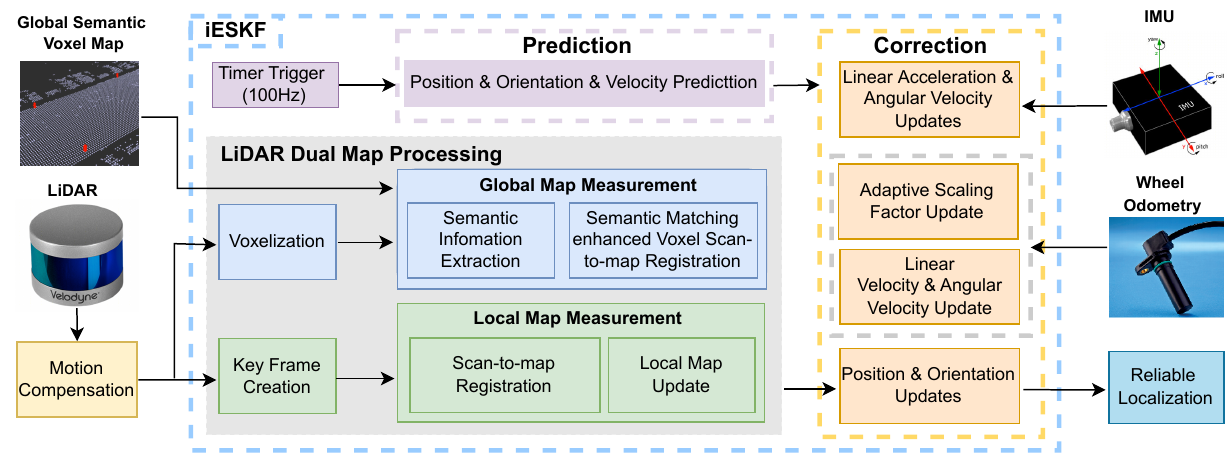}
    \caption{The diagram provides a detailed overview of our high-precision localization system architecture based on the \textbf{iESKF} filter, which integrates data from LiDAR, IMU, and wheel odometry, alongside a global voxel map enriched with semantic information for robust state estimation. The workflow is divided into four key modules: \textbf{LiDAR Dual Map Processing}, \textbf{Prediction}, and \textbf{Correction}.}
    \label{Fig.1}
    \vspace{-1.5 em}
\end{figure*}

% \begin{figure}[t]
%     \centering
%     \includegraphics[width=\linewidth]{Pictures/Architecture-LIWO-20250627-1.png}
%     \caption{The diagram provides a detailed overview of our high-precision localization system architecture based on the \textbf{iESKF} filter, which integrates data from LiDAR, IMU, and wheel odometry, alongside a global voxel map enriched with semantic information for robust state estimation. The workflow is divided into four key modules: \textbf{Inputs}, \textbf{LiDAR Dual Map Processing}, \textbf{Prediction}, and \textbf{Correction}.}
%     \label{Fig.1}
%     \vspace{-1 em}
% \end{figure}

As illustrated in Fig.~\ref{Fig.1}, our system framework comprises several key modules.
Sec.~\ref{sec:system_decri} introduces our tightly coupled iESKF-based localization backbone.
Sec.~\ref{sec:mapping} presents the voxel map construction and update pipeline with fused semantic information.
Sec.~\ref{sec:pred_model} presents a prediction model to estimate motion dynamics based on the current state.
Sec.~\ref{sec:lidar_dual_update} shows the integration of a global semantically-enhanced voxel map and a dynamically updated local map to achieve precise pose estimation and continuous local map refinement.
Sec.~\ref{sec:imu_correction} proposes the integration of IMU data to refine the predicted state by updating linear acceleration and angular velocity.
Sec.~\ref{sec:wheel_odometry_correction} shows the application of an adaptive scaling strategy to wheel odometry, where the constraints on linear and angular velocity are imposed to effectively enhance state estimation accuracy.

% As illustrated in Fig.~\ref{Fig.1}, our system framework is composed of several key modules. 
% In Sec.~\ref{sec:system_decri}, we introduce our localization backbone based on the tightly coupled iESKF framework.
% In Sec.~\ref{sec:mapping}, we outline the pipeline for voxel map construction and updating with semantic information fusion.
% In Sec.~\ref{sec:pred_model}, a constant velocity model is utilized to estimate motion dynamics based on the current state. 
% In Sec.~\ref{sec:lidar_dual_update}, this module integrates a global voxel map enriched with semantic information and a dynamically updated local map to enable precise pose estimation and continuous local map updates. 
% In Sec.~\ref{sec:imu_correction}, IMU data can be used to refine the predicted state by updating linear acceleration and angular velocity. 
% In Sec.~\ref{sec:wheel_odometry_correction}, by utilizing an adaptive scaling model, wheel odometry imposes linear and angular velocity constraints, effectively enhancing state estimation accuracy.

\subsection{System Description}
\label{sec:system_decri}

Our localization system, built on the mapping module, uses a tightly coupled iESKF-based multi-sensor fusion approach~\cite{9372856, 9697912} to enhance accuracy and reliability.

Firstly, 
we define the state vector \textbf{x} as
\begin{equation}
\label{eq:attn_state}
\resizebox{0.90\linewidth}{!}{
\(
\mathbf{x} \triangleq \begin{array}{l} 
\left[ \;\;
    ^{W}\mathbf{p}_{I}\;\;\;
    ^{W}\mathbf{v}_{I}\;\;\;
    ^{W}\mathbf{R}_{I}\;\;\;\;
    \mathbf{b}_{\mathbf{a}}\;\;\;\:\;
    \mathbf{b}_{\boldsymbol{\omega}}\;\;\;\;
    ^{B}\mathbf{a}\;\;\;\;\;
    ^{B}\boldsymbol{\omega} \right. \\ \left. \qquad\quad\;\;\;\;
    ^{v}\mathbf{S}\;\;\;\:\;
    ^{W}\mathbf{g}\;\;\;\;\;
    ^{I}\mathbf{R}_{L}\;\,\;
    ^{I}\mathbf{p}_{L}\;\,\;
    ^{I}\mathbf{R}_{B}\;\,\;\,
    ^{I}\mathbf{p}_{B} \;\; \right]
\end{array} \tag{1} 
\) 
}
\end{equation}
% \begin{equation}
% \label{eq:attn_state}
% \resizebox{0.85\linewidth}{!}{
% \(
% \mathbf{x} \triangleq \begin{bmatrix} \;\;
%     ^{W}\mathbf{p}_{I}\;\;
%     ^{W}\mathbf{v}_{I}\;\;
%     ^{W}\mathbf{R}_{I}\;\;
%     \mathbf{b}_{\mathbf{a}}\;\;
%     \mathbf{b}_{\boldsymbol{\omega}}\;\;
%     ^{B}\mathbf{a}\;\;
%     ^{B}\boldsymbol{\omega}\;\;
%     ^{v}\mathbf{S}\;\;
%     ^{W}\mathbf{g}\;\; \\
%     ^{I}\mathbf{R}_{L}\;\;
%     ^{I}\mathbf{p}_{L}\;\;
%     ^{I}\mathbf{R}_{B}\;\;
%     ^{I}\mathbf{p}_{B}
% \end{bmatrix} \tag{1} 
% \) 
% }
% \end{equation}
% \begin{equation}
% \label{eq:attn_state}
% \resizebox{0.80\linewidth}{!}{
% \(
% \mathbf{x} \triangleq \begin{bmatrix}
%     ^{W}\mathbf{p}_{I}\;\,
%     ^{W}\mathbf{v}_{I}\;\,
%     ^{W}\mathbf{R}_{I}\;\,
%     \mathbf{b}_{\mathbf{a}}\;\,
%     \mathbf{b}_{\boldsymbol{\omega}}\;\,
%     ^{B}\mathbf{a}\;\,
%     ^{B}\boldsymbol{\omega}\;\,
%     ^{v}\mathbf{S}\;\,
%     ^{W}\mathbf{g}\;\, \\
%     ^{I}\mathbf{R}_{L}\;\,
%     ^{I}\mathbf{p}_{L}\;\,
%     ^{I}\mathbf{R}_{B}\;\,
%     ^{I}\mathbf{p}_{B}
% \end{bmatrix}
% \)
% }
% \end{equation}
% \begin{equation}
% \label{eq:attn_state}
% \resizebox{0.9\linewidth}{!}{
% \(
% \mathbf{x} \triangleq \begin{bmatrix}
%     ^{W}\mathbf{R}_{I} &
%     ^{W}\mathbf{p}_{I} &
%     ^{W}\mathbf{v}_{I} &
%     ^{B}\boldsymbol{\omega} &
%     ^{B}\mathbf{a} &
%     \mathbf{b}_{\boldsymbol{\omega}} &
%     \mathbf{b}_{\mathbf{a}} &
%     \mathbf{S} &
%     ^{W}\mathbf{g} &
%     ^{I}\mathbf{R}_{L} &
%     ^{I}\mathbf{p}_{L} &
%     ^{I}\mathbf{R}_{B} &
%     ^{I}\mathbf{p}_{B}
% \end{bmatrix}
% \)
% }
% \end{equation}
where \( ^{W}\mathbf{p}_I \), \( ^{W}\mathbf{v}_I \), and \( ^{W}\mathbf{R}_I \) denote the IMU position, velocity and orientation in the world frame $W$. 
\( ^{B}\mathbf{a} \) and \( ^{B}\boldsymbol{\omega} \)  are the linear acceleration and angular velocity in the wheel odometry frame $B$, while \( \mathbf{b}_{\mathbf{a}} \) and \( \mathbf{b}_{\boldsymbol{\omega}} \) are the corresponding biases. 
\( ^{v}\mathbf{S} \) represents the adaptive scaling factor for linear velocity part of wheel odometry. \( ^{W}\mathbf{g} \) is the gravity vector in the world frame $W$. \( ^{I}\mathbf{R}_{L} \) and \( ^{I}\mathbf{p}_{L} \) are the extrinsic parameters between IMU frame $I$ and LiDAR frame $L$, respectively. \( ^{I}\mathbf{R}_{B} \) and \( ^{I}\mathbf{p}_{B} \) are the extrinsic parameters between IMU frame $I$ and wheel odometry frame $B$, respectively.

Using the notations \( \boxplus \) and \( \boxminus \) as defined in FAST-LIO \cite{9372856}, the state transition model in the Lie algebra space can be defined as:
\begin{equation}
    \label{eq:state_trans}
    \mathbf{x}_{i+1}=\mathbf{x}_{i} \boxplus \left(\Delta t \, \mathbf{f}\!\left(\mathbf{x}_{i}, \mathbf{w}_{i}\right)\right) \tag{2} 
\end{equation}
with the function \textbf{f} in forward process derived as
\begin{equation}
\resizebox{0.80\linewidth}{!}{
\(
\renewcommand{\arraystretch}{1.15} % 设置行距为默认的1.15倍
\mathbf{f}(\mathbf{x}, \mathbf{w}) \triangleq \begin{bmatrix}
    ^{W}\!\mathbf{v}_{I} + \frac{1}{2} \left( ^{W}\mathbf{R}_{I} \! ^{I}\mathbf{R}_{B} \! ^{B}\mathbf{a} + ^{W}\!\mathbf{g}\right)\Delta t \\
    ^{W}\mathbf{R}_{I} \! ^{I}\mathbf{R}_{B} \! ^{B}\mathbf{a} + ^{W}\!\mathbf{g} \\
    ^{I}\mathbf{R}_{B} \! ^{B}\boldsymbol{\omega} \\
    \mathbf{n}_{\mathbf{b a}} \\
    \mathbf{n}_{\mathbf{b} \boldsymbol{\omega}} \\
    \mathbf{n}_{\mathbf{a}} \\
    \mathbf{n}_{\boldsymbol{\omega}} \\
    % \mathbf{0}_{3 \times 1} \\
    % \mathbf{0}_{3 \times 1} \\
    % \mathbf{0}_{3 \times 1} \\
    % \mathbf{0}_{3 \times 1} \\
    % \mathbf{0}_{3 \times 1} \\
    \mathbf{0}_{18 \times 1}
\end{bmatrix}  \tag{3}
\)
}
\end{equation}

Within the system, IMU observation data is transformed from the IMU frame $I$ to the wheel odometry frame $B$ before being processed by the filter. Consequently, the IMU's angular velocity and linear acceleration measurement process noise, denoted as \( ^{B}\mathbf{w} \), is directly expressed in the wheel odometry frame $B$, as shown in \eqref{eq:noise}. Here, \( \mathbf{n}_{\mathbf{a}} \) and \( \mathbf{n}_{\boldsymbol{\omega}} \) denote the measurement noise of the IMU's linear acceleration and angular velocity, respectively, in the wheel odometry frame $B$, both of which can be modeled as Gaussian white noise. Meanwhile, the IMU measurement bias noise terms \( \mathbf{n}_{\mathbf{b a}} \) and \( \mathbf{n}_{\mathbf{b} \boldsymbol{\omega}} \), corresponding to the noise components of linear acceleration bias \( \mathbf{b}_{\mathbf{a}} \) and angular velocity bias \( \mathbf{b}_{\boldsymbol{\omega}} \) in the wheel odometry frame $B$, can be modeled as random walk processes using the following equations: 
\begin{equation*} 
\label{eq:noise}
\begin{aligned}
    ^{B}\mathbf{w} &\triangleq \begin{bmatrix}
        \mathbf{n}_{\mathbf{a}} &
        \mathbf{n}_{\boldsymbol{\omega}} &
        \mathbf{n}_{\mathbf{b a}} &
        \mathbf{n}_{\mathbf{b} \boldsymbol{\omega}}
    \end{bmatrix} \\
    \mathbf{n}_a \sim&\; \mathcal{N}(0, \sigma_a^2), \quad \mathbf{n}_\omega \sim \mathcal{N}(0, \sigma_\omega^2). \\
    \mathbf{n}_{ba} \sim&\; \mathcal{N}(0, \sigma_{b_a}^2), \quad \mathbf{n}_{b\omega} \sim \mathcal{N}(0, \sigma_{b_\omega}^2).
\end{aligned} \tag{4} 
\end{equation*}
% \begin{equation}
%  ^{B}\mathbf{w} \triangleq \begin{bmatrix}
%     \mathbf{n}_{\mathbf{a}} &
%     \mathbf{n}_{\boldsymbol{\omega}} &
%     \mathbf{n}_{\mathbf{b a}} &
%     \mathbf{n}_{\mathbf{b} \boldsymbol{\omega}}
% \end{bmatrix}
% \end{equation}
% \begin{equation}
%     \mathbf{n}_a \sim \mathcal{N}(0, \sigma_a^2), \quad \mathbf{n}_\omega \sim \mathcal{N}(0, \sigma_\omega^2).
% \end{equation}
% \begin{equation}
%     \mathbf{n}_{ba} \sim \mathcal{N}(0, \sigma_{b_a}^2), \quad \mathbf{n}_{b\omega} \sim \mathcal{N}(0, \sigma_{b_\omega}^2).
% \end{equation}

\subsection{Map Construction and Update}
\label{sec:mapping}
% High-precision LiDAR mapping enables accurate localization in complex and dynamic environments, while optimizing point cloud storage formats and integrating semantic information significantly improve map management efficiency and localization precision.
% The mapping module adopts a dual-map architecture, consisting of a global map and a local dynamic map, each tailored to fulfill distinct roles. The global map serves as a long-term stable reference framework, ensuring the consistency and reliability of global localization. The local dynamic map focuses on quickly adapting to changes in dynamic environments, providing immediate short-term real-time localization support.
The mapping module utilizes a dual-map architecture, as demonstrated in Fig.~\ref{Fig.1}. The global map provides a stable long-term reference for consistent and reliable localization, while the local dynamic map focuses on rapid adaptation to environmental changes, offering real-time support in dynamic scenarios. This hierarchical structure combines coarse alignment via the global map and fine alignment with the local map, ensuring accurate, efficient, and reliable localization across diverse conditions.

\noindent\textbf{Map Construction.}
In constructing pre-existing global maps, Fast-LIO2 \cite{9697912} incorporates loop closure and GPS factors for back-end optimization (inspired by the LIO-SAM \cite{9341176} framework), ensuring precise alignment and high consistency of point cloud data. By detecting and removing dynamic elements while retaining only static point cloud data, it effectively prevents localization drift caused by environmental changes or moving objects. 
For local dynamic maps, key frames capturing core point cloud data are generated based on significant environmental changes to reduce redundancy and improve update efficiency. Then, corrected LiDAR poses are used to integrate multi-frame key frames, incrementally updating the map based on map size and distance thresholds to quickly adapt to environmental changes and ensure short-term localization accuracy.
% In the process of constructing global map, the integration of the Fast-LIO2 algorithm \cite{9697912} with inertial navigation system (INS) data ensures precise alignment of point cloud data, maintaining overall consistency and high accuracy. 
% Additionally, dynamic elements are detected and removed, retaining only static point cloud data. This approach effectively prevents localization drift caused by environmental changes or interference from moving objects, significantly enhancing the stability and reliability of the global map. 
% For the local dynamic map, corrected LiDAR poses are utilized to integrate multi-frame real-time point cloud data, dynamically updating the map by adding new point cloud data and removing outdated data based on the map's size and predefined distance thresholds. This mechanism enables the system to quickly adapt to changes in dynamic environments, such as pedestrians, vehicles, or other moving objects, ensuring the local map consistently maintains high accuracy to support short-term precise localization.

\noindent\textbf{Voxel-Based Storage.}
A voxel-based point cloud storage format (referencing VSCE in iG-LIO \cite{10380742}) divides the space into fixed 0.5-meter voxel grids, with each voxel retaining a representative point, characterized by the mean coordinates and distribution variance of points within the voxel, to significantly reduce data redundancy, improve registration efficiency, and support fast neighborhood queries and dynamic updates. 
% During the construction of both global and local maps, the data storage format is optimized to enhance system performance. Specifically, a voxel-based point cloud storage format, similar to the VSCE approach in iG-LIO \cite{10380742}, is adopted to divide the three-dimensional space into fixed-size voxel grids (0.5 meters), significantly reducing data redundancy. Each voxel retains only a single representative point, characterized by the mean coordinates and distribution variance of points within the voxel. This approach not only minimizes the data volume but also greatly improves the efficiency of point cloud registration and retrieval. 
% Particularly, in our scenario, when handling large-scale point cloud datasets, this method drastically reduces computational resource consumption. 
% Moreover, the voxel format inherently supports spatial indexing, enabling fast neighborhood queries and dynamic updates. This capability makes the system highly suitable for real-time map construction and maintenance in complex dynamic environments.

\noindent\textbf{Global Semantic Information Integration.}
Building on dynamic object removal and voxelization, semantic features are extracted from the voxelized global map using Algorithm \ref{alg:semantic_info_extract}. This enables the system to leverages high-level semantic environmental information for more reliable global point cloud registration and alignment. Meanwhile, to enhance the robustness and accuracy of point cloud registration, especially in handling degenerate scenarios or noisy point clouds, we introduced specific constraint strategies for different semantic feature voxels. For cylinder semantic feature voxels, a minimum eigenvalue constraint is applied to improve numerical stability and prevent optimization failure caused by degenerate directions. For plane and other semantic feature voxels, a plane eigenvalue constraint is introduced to enhance registration accuracy in planar regions and mitigate errors caused by planar feature degeneration.
% \textcolor{red}{\st{, especially in environments with repetitive or ambiguous geometric patterns. Semantic-based matching strengthens the distinction between dynamic and static elements, deepening environmental understanding. This fusion of voxel-based storage and semantics optimizes map management efficiency while significantly improving localization accuracy, robustness, and adaptability in dynamic, complex environments.}}

\vspace{-0.5 em}
\begin{algorithm}[t]
\caption{Voxel-based Semantic Information Extraction and Covariance Refinement}
\label{alg:semantic_info_extract}
\begin{algorithmic}[1]
\item \textbf{Input:} Voxel map $\mathcal{M}$ with mean $\mu_g$ and covariance $\Sigma_g$ for each grid, angle threshold $\theta_{thc}$ for \textbf{cylinder}, angle threshold $\theta_{thp}$ for \textbf{plane}. \\
\textbf{Output:} Voxel map $\mathcal{M}$ with semantic information \textbf{cylinder}, \textbf{plane}, or \textbf{other} for each grid.
\FOR{each grid $g \in \mathcal{M}$}
    \STATE Perform SVD on $\Sigma_g$: $\Sigma_g = \mathbf{U} \mathbf{\Sigma} \mathbf{V}^\top$; 
    \STATE Extract the singular values $\sigma_1$, $\sigma_2$, $\sigma_3$ from $\mathbf{\Sigma}$, along with their corresponding singular vectors $\mathbf{u}_1$, $\mathbf{u}_2$, $\mathbf{u}_3$ from $\mathbf{U}$, ensuring the singular values satisfy the condition $\sigma_1 \geq \sigma_2 \geq \sigma_3$;
    \STATE Set grid $g$ as \textbf{other} by default;
    \IF{$\sigma_1 \gg \sigma_2$ \AND $\sigma_2 \approx \sigma_3$} 
        \STATE Compute angle $\theta = \arccos\left(\frac{\mathbf{u}_1 \cdot \mathbf{z}}{\|\mathbf{u}_1\| \|\mathbf{z}\|}\right)$, where $\mathbf{z}$ is the unit vector of the z-axis;
        \IF{$\theta < \theta_{thc}$}
            \STATE Classify grid $g$ as \textbf{cylinder};
        \ENDIF
    \ELSIF{$\sigma_1 \approx \sigma_2$ \AND $\sigma_3 \ll \sigma_1$} 
        \STATE Compute angle $\theta = \arccos\left(\frac{\mathbf{u}_3 \cdot \mathbf{z}}{\|\mathbf{u}_3\| \|\mathbf{z}\|}\right)$, where $\mathbf{z}$ is the unit vector of the z-axis;
        \IF{$\theta < \theta_{thp}$}
            \STATE Classify grid $g$ as \textbf{plane};
        \ENDIF
    \ENDIF
    \IF{grid $g$ is \textbf{cylinder}}
        \STATE Compute $\sigma_i = \max(\sigma_i, 1\text{e-3})$ for the singular values $\sigma_1$, $\sigma_2$, $\sigma_3$
        \STATE Set the diagonal matrix $\mathbf{D} = \text{diag}(\sigma_1,  \sigma_2, \sigma_3)$;
    \ELSIF{grid $g$ is \textbf{plane} \OR grid $g$ is \textbf{other}}
        \STATE Set the diagonal matrix $\mathbf{D} = \text{diag}(1, 1, 1\text{e-3})$;
    \ENDIF
    \STATE Update the covariance of grid $g$: $\Sigma_g = \mathbf{U} \mathbf{D} \mathbf{V}^\top$;
\ENDFOR

\end{algorithmic}
\end{algorithm}

\subsection{Model Prediction}
\label{sec:pred_model}

% In the case of sensor failures, such as data interruptions from the IMU, we introduce a 100 Hz timer and switch to a constant velocity model for prediction. This method ensures that even in the absence of IMU data, the localization system does not get stuck waiting for data synchronization or immediately diverge. The system assumes that the linear and angular velocity states remain constant between two consecutive LiDAR frames, allowing the entire positioning pipeline to degrade into a pure LiDAR-based localization method using the constant velocity model. The prediction model, as applied during the filter prediction phase, can be expressed as:

% In the prediction module, we utilize a constant velocity model to estimate the system state, assuming that linear and angular velocities remain constant between two observation frames. A 100 Hz timer is implemented to periodically trigger state predictions. Compared to IMU pre-integration-based prediction methods, the constant velocity model offers notable advantages in scenarios involving IMU failures or data interruptions. IMU pre-integration relies on continuous high-frequency inertial measurements, where data interruptions can result in unstable predictions, decreased localization accuracy, or even filter divergence. In contrast, the constant velocity model is independent of IMU data continuity and can maintain stable state predictions even in the absence of IMU measurements, effectively preventing system stagnation or immediate divergence caused by synchronization delays. 

In our prediction model, the updated state \(\mathbf{\bar{x}}_i\) at time step \( i \) can be propagated to the next time step \( i+1 \) using the state transition model described in \eqref{eq:state_trans}, with \( \mathbf{w}_i \) set to zero:
\begin{equation}
    \mathbf{\hat{x}}_{i+1} = \mathbf{\bar{x}}_i \boxplus \left(\Delta t \, \mathbf{f}\!\left(\mathbf{\bar{x}}_{i}, 0\right)\right) \tag{5}
\end{equation}
In greater detail, the state can be propagated using the following equations:
\begin{gather*}
    ^{W}\mathbf{\hat{p}}_{I}\!\!\!\mathrel{}^{i+1} = ^{W}\!\!\mathbf{\bar{p}}_{I}\!\!\!\mathrel{}^{i} + ^{W}\!\mathbf{\bar{v}}_{I}\!\!\!\mathrel{}^{i} \Delta t + \frac{1}{2} \left( ^{W}\mathbf{\bar{R}}_{I}\!\!\!\mathrel{}^{i} \! ^{I}\mathbf{\bar{R}}_{B}\!\!\!\mathrel{}^{i} \! ^{B}\mathbf{\bar{a}}\!\!\mathrel{}^{i} + ^{W}\!\mathbf{\bar{g}}\!\!\mathrel{}^{i}\right) \Delta t^{2} \\
    ^{W}\!\mathbf{\hat{v}}_{I}\!\!\!\mathrel{}^{i+1} = ^{W}\!\!\mathbf{\bar{v}}_{I}\!\!\!\mathrel{}^{i} + \left(^{W}\mathbf{\bar{R}}_{I}\!\!\!\mathrel{}^{i} \! ^{I}\mathbf{\bar{R}}_{B}\!\!\!\mathrel{}^{i} \! ^{B}\mathbf{\bar{a}}\!\!\mathrel{}^{i} + ^{W}\!\mathbf{\bar{g}}\!\!\mathrel{}^{i}\right) \Delta t \\
    ^{W}\mathbf{\hat{R}}_{I}\!\!\!\mathrel{}^{i+1} = ^{W}\!\!\mathbf{\bar{R}}_{I}\!\!\!\mathrel{}^{i} \mathbf{Exp}\left( ^{I}\mathbf{\bar{R}}_{B}\!\!\!\mathrel{}^{i} \! ^{B}\boldsymbol{\bar{\omega}}\!\!\mathrel{}^{i}\,\Delta t\right) \tag{6}
\end{gather*}
Rather than relying on raw IMU measurements, state propagation can be achieved by leveraging estimated linear acceleration and angular velocity and assuming these values remain constant between observation frames. In contrast to pure IMU pre-integration methods, which depend on continuous high-frequency inertial data and are vulnerable to filter instability or divergence during data interruptions, this approach demonstrates strong resilience against IMU failures, data gaps, and synchronization delays.

Then, the propagation of covariance \(\mathbf{\bar{P}}_i\) from time step \( i \) to \( i+1 \) can be implemented as follows:
\begin{equation}
    \mathbf{\hat{P}}_{i+1} = \mathbf{F}_{\mathbf{x}_i} \mathbf{\bar{P}}_i \mathbf{F}_{\mathbf{x}_i}^\top + \mathbf{F}_{\mathbf{w}_i} \mathbf{Q}_i \mathbf{F}_{\mathbf{w}_i}^\top \tag{7}
\end{equation}
where \( \mathbf{Q}_i \) is the covariance of the IMU measurement process noise \(\mathbf{w}_i\), and the matrices \( \mathbf{F}_{\mathbf{\bar{x}}_i} \) and \( \mathbf{F}_{\mathbf{w}_i} \) can be calculated as below:
\begin{gather*}
    \mathbf{F}_{\mathbf{x}_i} = \left. \frac{\partial (\mathbf{x}_{i+1} \boxminus \mathbf{\hat{x}}_{i+1})}{\partial \delta \mathbf{x}_i} \right|_{\delta \mathbf{x}_i=0, \mathbf{w}_i=0} \\
    \mathbf{F}_{\mathbf{w}_i} = \left. \frac{\partial (\mathbf{x}_{i+1} \boxminus \mathbf{\hat{x}}_{i+1})}{\partial \mathbf{w}_i} \right|_{\delta \mathbf{x}_i=0, \mathbf{w}_i=0} \\
    \text{where} \quad \delta \mathbf{x}_i = \mathbf{x}_i \boxminus \mathbf{\hat{x}}_i \tag{8}
\end{gather*}

\noindent\subsection{LiDAR Dual Map Processing}
\label{sec:lidar_dual_update}

The LiDAR dual-map processing module consists of global map measurement and local map measurement, which handle global and local point cloud data respectively, working together to achieve precise and efficient localization. First, we define some symbols. \( ^{L_{i}}\mathbf{p}_{j} \) and \( ^{L_{i}}{\mathbf n}_j \) represent the LiDAR point $j$ in current $i$-th frame of point cloud and its noise, and we assume that this noise is affected by zero-mean Gaussian white noise. \( ^\mathcal{M}\mathbf{T}_{I}\!\!\!\mathrel{}^{i} \) is the $i$-th estimate of transformation between the global map frame \( \mathcal{M} \) and the IMU frame \( \mathbf{I} \). \( ^{I}\mathbf{T}_{L}\!\!\!\mathrel{}^{i} \) is the $i$-th estimate of transformation between the IMU frame \( \mathbf{I} \) and the LiDAR frame \( \mathbf{L} \).

\noindent\textbf{Global Map Measurement}. 
The global pose can be estimated by matching the current point cloud with the global voxel-based prior map \( \mathcal{M}_{global} \), ensuring long-term localization stability. First, the current point cloud is voxelized, dividing it into 0.5-meter voxel units to effectively reduce data volume while preserving the spatial structure of the environment, thereby improving computational efficiency. During the voxel-based scan-to-map registration phase, the voxelized current point cloud is aligned with the global map using the K-nearest neighbor voxel search method and GICP registration \cite{Segal2009GeneralizedICP}, as described below, for pose estimation: 
\vspace{-1mm} % 减少公式上方的空行
\begin{equation}
\label{eq:gicp}
\resizebox{1.00\linewidth}{!}{
\(
    ^{\mathcal{M}_{global}}\mathbf{R}_{j} \! \left(\!\mathbf{x}_{i}, \!^{L_{i}}\mathbf{p}_{j}, \!^{L_{i}}\mathbf{n}_{j}\! \right) = \mathbf{Vox}\!\!\left( ^\mathcal{M}\mathbf{q}_{j} \right) - \mathbf{Vox}\!\!\left(\!{^\mathcal{M}\mathbf{T}_{I}}\!\!\!\mathrel{}^{i} \!{^{I}\mathbf{T}_{L}}\!\!\!\mathrel{}^{i} \!\! \left(\!^{L_{i}}\mathbf{p}_{j}\!+\! ^{L_{i}}\mathbf{n}_{j} \!\right)\right) \tag{9}
\)
}
\end{equation}
where \( \mathbf{Vox} \) represents the voxelization operation applied to a point in the map, returning the mean value of the corresponding voxel. \( ^\mathcal{M}\mathbf{q}_{j} \) is the associated point in the global map.

Since voxels with semantic feature labels are extracted from the pre-existing map using Algorithm \ref{alg:semantic_info_extract}, different weights are assigned to these voxels during the computation of constraints. Specifically, \(w_\text{cylinder}\), \(w_\text{plane}\), and \(w_\text{other}\) correspond to voxels with cylinder, plane, and other semantic features, respectively, and the weights satisfy the relationship \(w_\text{cylinder} > w_\text{plane} > w_\text{other}\). 
Then, based on the semantic label of the \textbf{voxel} that the LiDAR point $j$ in the current $i$-th frame of the point cloud belongs to, the weight \(w_j\) for this point $j$ can be determined using an indicator function $\mathbb{I}$, as shown below:
\begin{gather*}
\label{eq:weight_calc}
\begin{aligned}
    w_j = \,\,
    &w_\text{cylinder} \,\, \cdot \, \mathbb{I}\,\{\mathbf{voxel} \in cylinder\} \\ 
    &+  w_\text{plane} \, \cdot \, \mathbb{I}\,\{\mathbf{voxel} \in plane\}     \\ 
    &+  w_\text{other} \, \cdot \, \mathbb{I}\,\{\mathbf{voxel} \in other\} \\
    \text{where } &\mathbb{I}\{\text{condition}\} =
        \begin{cases} 
        1, & \text{if condition is true}, \\
        0, & \text{otherwise}.
        \end{cases}
\end{aligned} \tag{10}
\end{gather*}

%%%%%%%%%%%%%%%%%%%%%%%%%%%%%%%
\begin{figure}[t]
    \centering
    \includegraphics[width=0.85\linewidth]{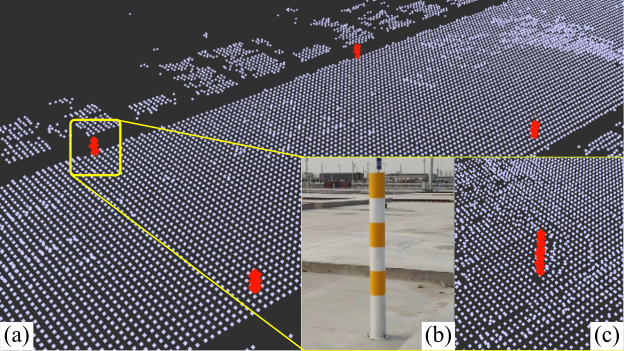} % 插入图片
    % 总标题
    \caption{
    % Pillars on the point cloud map. 
    (a) Example of deployment of pillars. Pillars are strategically deployed along both sides of the lane, with a spacing interval of 35\,m between each pillar. (b) The pillar features a diameter of at least 20\,cm and a minimum height of 1.5\,m. (c) The point cloud representation of the pillars is highlighted in red on the map, while the background point cloud is shown in white. }
    \label{fig:pillar}
    \vspace{-1 em}
\end{figure}

\noindent\textbf{Local Map Measurement}. 
By concentrating on the immediate vicinity, the local map enables faster and more efficient scan-to-map registration, reducing the computational load of the global map. Meanwhile, the local map is dynamically updated to capture changes like moving objects or temporary obstacles, enhancing robustness in cluttered environments.
% The smaller area of the local map ensures faster matching, suitable for real-time applications. 
The current point cloud is matched to this dynamically updated local voxel map \( \mathcal{M}_{local} \) for relative pose calculation using the KNN search algorithm and point-to-plane registration, as detailed below:   
% \vspace{-1mm} % 减少公式上方的空行
\begin{equation}
\label{eq:p2s}
\resizebox{1.00\linewidth}{!}{
\(
    ^{\mathcal{M}_{local}}\mathbf{R}_{j} \! \left(\!\mathbf{x}_{i}, \!^{L_{i}}\mathbf{p}_{j}, \!^{L_{i}}\mathbf{n}_{j}\! \right) = {{^\mathcal{M}\mathbf{u}_{j}}\!^{T}}\! \left(\!{^\mathcal{M}\mathbf{T}_{I}}\!\!\!\mathrel{}^{i} \!{^{I}\mathbf{T}_{L}}\!\!\!\mathrel{}^{i} \! \left(\!^{L_{i}}\mathbf{p}_{j} + ^{L_{i}}\mathbf{n}_{j} \!\right) - ^\mathcal{M}\mathbf{q}_{j}\!\right) \tag{11}
\)
}
\end{equation}
where \( ^\mathcal{M}\mathbf{u}_j \) is the normal vector of the associated plane fitted using neighboring points of \( ^{L_{i}}\mathbf{p}_{j} \) in the local map. \( ^\mathcal{M}\mathbf{q}_{j} \) is another point on the associated fitted plane in the local map.

\noindent\textbf{Overall Map Constraints}.
Rather than directly registering point clouds with the map, we integrate feature alignment into the iESKF measurement update through corresponding constraints. This process is expressed by the following equation: 
\begin{equation}
\label{eq:combine_constraints}
    \mathbf{0}\!=w_j \cdot \!^{\mathcal{M}_{global}}\mathbf{R}_{j} \! \left(\!\mathbf{x}_{i}, \!^{L_{i}}\mathbf{p}_{j}, \!^{L_{i}}\mathbf{n}_{j}\!\right)\! + \!^{\mathcal{M}_{local}}\mathbf{R}_{j} \!\left(\!\mathbf{x}_{i}, \!^{L_{i}}\mathbf{p}_{j}, \!^{L_{i}}\mathbf{n}_{j}\!\right) \tag{12}
\end{equation} 
This approach combines high-accuracy registration of the global map \( \mathcal{M}_{\text{global}} \) with the real-time matching of the local map \( \mathcal{M}_{\text{local}} \) within a single frame. The weights \(w_j\) are calculated using \eqref{eq:weight_calc} and assigned to voxels representing cylinders, planes, and other semantic features.

% This approach strikes a balance between precision and efficiency while maintaining robustness against noise and environmental changes.

% To accelerate computation, a mathematically equivalent formula for Kalman gain from FAST-LIO \cite{9372856} is adopted, reducing complexity from observation to state dimension, ensuring efficient measurement updates even with large LiDAR point cloud observations.

%%%%%%%%%%%%%%%%%%%%%%%%%%%%%%%
\begin{figure}[t]
    \centering
    % 上方图片
    \begin{subfigure}[b]{0.8\linewidth} % 上方图片宽度占90%
        \centering
        \includegraphics[width=\linewidth]{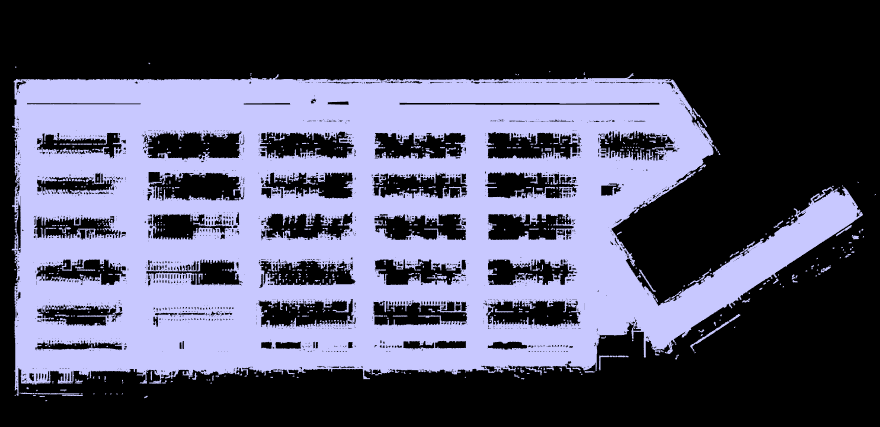} % 插入图片
        \caption{} % 子图标题
        \label{fig:top} % 子图标签
    \end{subfigure}
    
    % 下方图片
    \vspace{0.5em} % 上下图片之间的垂直间距
    \begin{subfigure}[b]{0.8\linewidth} % 下方图片宽度占90%
        \centering
        \includegraphics[width=\linewidth]{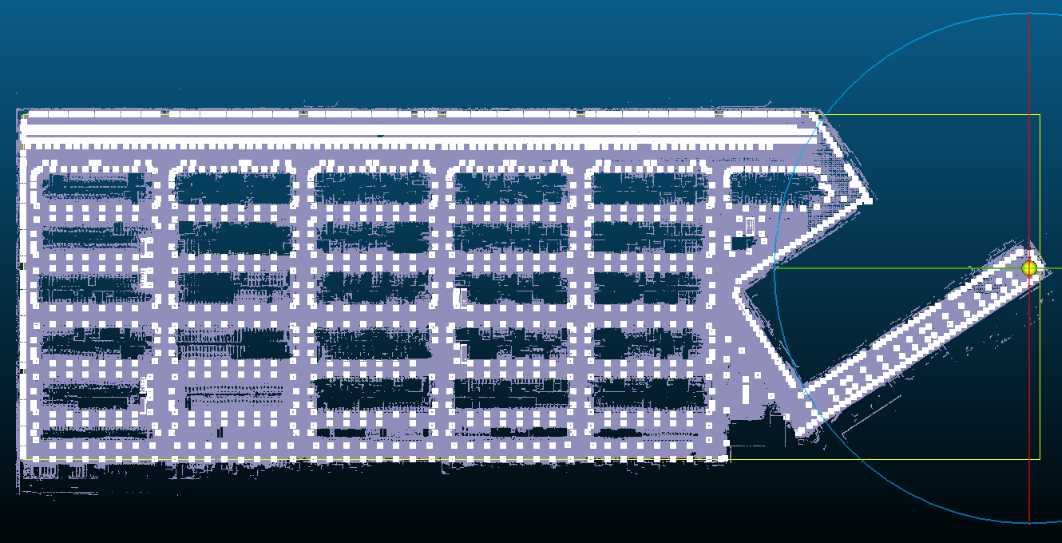} % 插入图片
        \caption{} % 子图标题
        \label{fig:bottom-2} % 子图标签
    \end{subfigure}
    
    % 总标题
    \caption{Global voxelized point cloud map with semantic information. (a) The generated point cloud map spans an area of 1538\,m $\times$ 596\,m. (b) The global semantic point cloud highlights pillar points, which are projected onto the global map with an enlarged view to emphasize their positions. The pillar points are represented in white, while the background global point cloud map is displayed in light purple. }
    \label{fig:maps}
    \vspace{-1em}
\end{figure}
%%%%%%%%%%%%%%%%%%%%%%%%%%%%%%%

\subsection{IMU-based Correction}
\label{sec:imu_correction}

The IMU is an indispensable sensor in the system, providing high-frequency linear acceleration and angular velocity data. In this system, we treat IMU data as a measurement for filter correction in the wheel odometry frame $B$.
% \begin{gather*}
%     ^{B}\boldsymbol{\omega} = ^{I}\!{\mathbf{R}_{B}^{-1}} \, ^{I}\boldsymbol{\omega}_{m} - \mathbf{n}_{\boldsymbol{\omega}} - \mathbf{b}_{\boldsymbol{\omega}} \\
%     ^{B}\mathbf{a} = ^{I}\!{\mathbf{R}_{B}^{-1}} \, ^{I}\mathbf{a}_{m} - \mathbf{n}_{\mathbf{a}} - \mathbf{b}_{\mathbf{a}} \tag{12}
% \end{gather*}
Then, for the $i$-th frame of IMU measurement data, we can denote the constraints as:
\begin{gather*}
\begin{aligned}
    &\mathbf{R}\left(\mathbf{x}_{i}, ^{I}\!\boldsymbol{\omega}^{i}_{m}, \mathbf{n}_{\boldsymbol{\omega}}^{i} \right) = \left(^{I}{\mathbf{R}_{B}^{i}}\right)^{-1}  \, ^{I}\boldsymbol{\omega}^{i}_{m} - \mathbf{n}_{\boldsymbol{\omega}}^{i} - \mathbf{b}_{\boldsymbol{\omega}}^{i} - ^{B}\!\boldsymbol{\omega}^{i}  \\
    &\mathbf{R}\left(\mathbf{x}_{i}, ^{I}\!\mathbf{a}^{i}_{m}, \,\mathbf{n}_{\mathbf{a}}^{i} \right) \,= \left(^{I}{\mathbf{R}_{B}^{i}}\right)^{-1} \, ^{I}\mathbf{a}^{i}_{m} \,- \mathbf{n}_{\mathbf{a}}^{i} \,- \mathbf{b}_{\mathbf{a}}^{i} \,- ^{B}\!\mathbf{a}^{i} 
\end{aligned}\tag{13}
\end{gather*}
where \( ^{I}\boldsymbol{\omega}^{i}_{m} \) and \( ^{I}\mathbf{a}^{i}_{m} \) are the $i$-th frame of measurements of IMU angular velocity and linear acceleration.

These IMU constraints can be integrated in the following form for the observation update:
\begin{equation}
    \mathbf{0} = \mathbf{R}\left(\mathbf{x}_{i}, ^{I}\!\boldsymbol{\omega}^{i}_{m}, \mathbf{n}_{\boldsymbol{\omega}}^{i} \right) + \mathbf{R}\left(\mathbf{x}_{i}, ^{I}\!\mathbf{a}^{i}_{m}, \mathbf{n}_{\mathbf{a}}^{i} \right)    \tag{14}
\end{equation}

\subsection{Wheel Odometry-based Correction with An Adaptive Scaling Model}
\label{sec:wheel_odometry_correction}

Wheel odometry provides low-drift measurements, complementing LiDAR and IMU in scenarios with poor visibility, dynamic obstacles, or high-frequency vibrations, and enhances robustness in low-feature environments.
% By tightly coupling wheel odometry with LiDAR and IMU, the system achieves high localization accuracy and adaptability. 
In addition, to address challenges like tire slippage on complex terrains, a 3D adaptive scaling factor model dynamically adjusts linear velocity observation weights, compensating errors and improving robustness and precision in dynamic and diverse terrains. 
% \begin{gather*}
%     ^{W}\mathbf{v}_{I} = ^{W}\!{\mathbf{R}_{I}} \left(^{I}\!\mathbf{R}_{B} \, ^{B}\mathbf{v}_{m} \odot \, ^{v}\mathbf{S} - [\mathbf{^{I}\mathbf{R}_{B} \, ^{B}\boldsymbol{\omega}}]_\times \, ^{I}\mathbf{t}_{B}\right) \\
%     ^{B}\boldsymbol{\omega} = ^{B}\!\boldsymbol{\omega}_{m} \tag{15}
% \end{gather*}
Then, upon receiving the $i$-th frame of wheel odometry measurement data, the constraints for the filter observation update can be expressed as:
\begin{gather*}
\begin{aligned}
    \mathbf{R}\left(\mathbf{x}_{i}, ^{B}\!\mathbf{v}_{m}\!\!\!\mathrel{}^{i}, \mathbf{0} \right) = &^{W}\!\!{\mathbf{R}_{I}}\!\!\!\mathrel{}^{i} {^{I}\mathbf{R}_{B}\!\!\!\mathrel{}^{i}} {^{B}\!\mathbf{v}_{m}\!\!\!\mathrel{}^{i}} \odot ^{v}\!\mathbf{S}\!\mathrel{}^{i} \\
    &- ^{W}\!\!{\mathbf{R}_{I}}\!\!\!\mathrel{}^{i}[\mathbf{^{I}\mathbf{R}_{B}\!\!\!\mathrel{}^{i} \, ^{B}\boldsymbol{\omega}\!\mathrel{}^{i}}]_\times \, ^{I}\mathbf{t}_{B}\!\!\!\mathrel{}^{i} \\
    &- ^{W}\!\mathbf{v}_{I}\!\!\!\mathrel{}^{i}\\
    \mathbf{R}\left(\mathbf{x}_{i}, ^{B}\!\boldsymbol{\omega}^{i}_{m}, \mathbf{0} \right) = &^{B}\!\boldsymbol{\omega}_{m}\!\!\!\mathrel{}^{i} - ^{B}\boldsymbol{\omega}\!\mathrel{}^{i} 
\end{aligned} \tag{15}
\end{gather*}
where \( ^{I}\boldsymbol{\omega}^{i}_{m} \) and \( ^{B}\mathbf{v}^{i}_{m} \) are the $i$-th frame of the measured angular velocity and linear velocity from the wheel odometry. The operator \( \odot \) denotes the element-wise product.

These constraints for wheel odometry can be integrated in the following form for the observation update:
\begin{equation}
    \mathbf{0} = \mathbf{R}\left(\mathbf{x}_{i}, ^{I}\!\boldsymbol{\omega}^{i}, \mathbf{n}_{\boldsymbol{\omega}}^{i} \right) + \mathbf{R}\left(\mathbf{x}_{i}, ^{I}\!\mathbf{a}^{i}, \mathbf{n}_{\mathbf{a}}^{i} \right)    \tag{16}
\end{equation}

% \subsection{Implement Details}
% \label{sec:details_imple}

% \textbf{LiDAR Motion Compensation}. When the LiDAR is in motion, point cloud data can become distorted due to sampling at different poses. To address this, we adopt a motion compensation strategy similar to Fast-LIO \cite{9372856}, leveraging filter state information. By maintaining a history of filter states over a period, we interpolate this data to estimate the precise state at the moment each point was captured. These points are then projected to a common reference frame, such as the end of the scan. This approach effectively compensates for motion distortions, ensuring coherent and accurate point cloud representation.

\section{EXPERIMENTS}

\begin{figure}[t]
    \centering
    \includegraphics[width=0.85\linewidth]{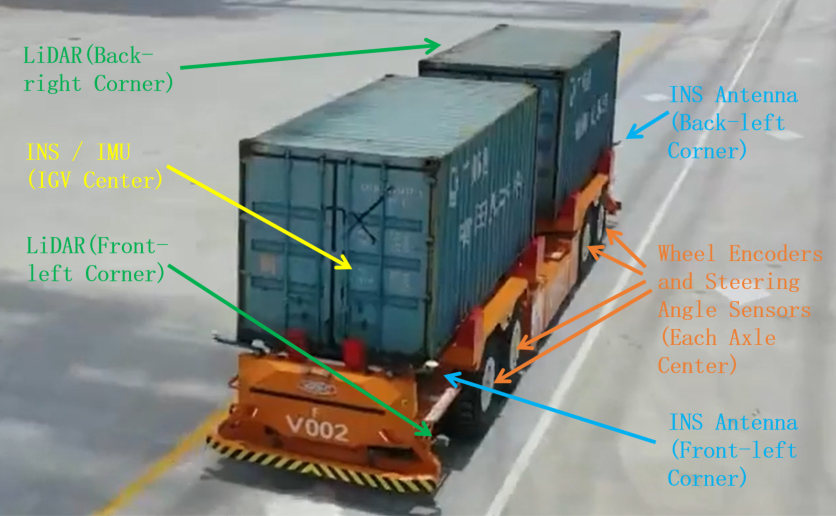}
    \caption{
    %The IGV has dimensions of 15\,m$\times$3\,m$\times$1.7\,m and is furnished with two 16-line LiDARs, \textcolor{red}{four wheel encoders, four steering angle sensors} and a INS (with two antennas and a 6-axis IMU).
    The IGV measures 15\,m$\times$3\,m$\times$1.7\,m and is equipped with two 16-line LiDARs, four wheel encoders, four steering angle sensors and an INS (featuring two antennas and a 6-axis IMU).
    }
    \label{fig:igv}
    \vspace{-1.5 em}
\end{figure}

% \begin{figure}[!t]
%     \centering
%     % 上方左侧图片
%     \begin{subfigure}[b]{0.225\textwidth} % 左侧图片宽度占45%
%         \centering
%         \includegraphics[height=4cm,keepaspectratio]{Pictures/pillar/pillar.png} % 插入图片
%         \caption{} % 子图标题
%         \label{fig:top-left} % 子图标签
%     \end{subfigure}
%     \hfill % 两张图片之间的水平间距
%     % 上方右侧图片
%     \begin{subfigure}[b]{0.225\textwidth} % 右侧图片宽度占45%
%         \centering
%         \includegraphics[height=4cm,keepaspectratio]{Pictures/pillar/pillar_on_map_zoom_in_4.png} % 插入图片
%         \caption{} % 子图标题
%         \label{fig:top-right} % 子图标签
%     \end{subfigure}
    
%     % 下方图片
%     \vspace{1 em} % 上下图片之间的垂直间距
%     \begin{subfigure}[b]{0.455\textwidth} % 下方图片宽度占90%
%         \centering
%         \includegraphics[width=\textwidth]{Pictures/pillar/pillar_on_map_zoom_in_5.png} % 插入图片
%         \caption{} % 子图标题
%         \label{fig:bottom-1} % 子图标签
%     \end{subfigure}
    
%     % 总标题
%     \caption{Pillars on the point cloud map. (a) The pillar, with a diameter of no less than 20\,cm and a height of at least 1.5\,m. (b) The point cloud representation of the pillars is highlighted in red on the map, while the background point cloud is shown in white. (c) Example of deployment of pillars. Pillars are strategically deployed along both sides of the lane, with a spacing interval of 35\,m between each pillar. }
%     \label{fig:pillar}
%     \vspace{-1.5 em}
% \end{figure}

To evaluate the performance of the proposed method, we have gathered real-world data from a typical dynamic and open port terminal to build a dataset for testing. Details of the experimental setup are provided in Sec.~\ref{sec:exp_set}, and the localization results are discussed in Sec.~\ref{sec:loc_result}.

\begin{figure}[t]
    \centering
   \subfloat[\label{fig:top}]{%
       \includegraphics[width=\linewidth]{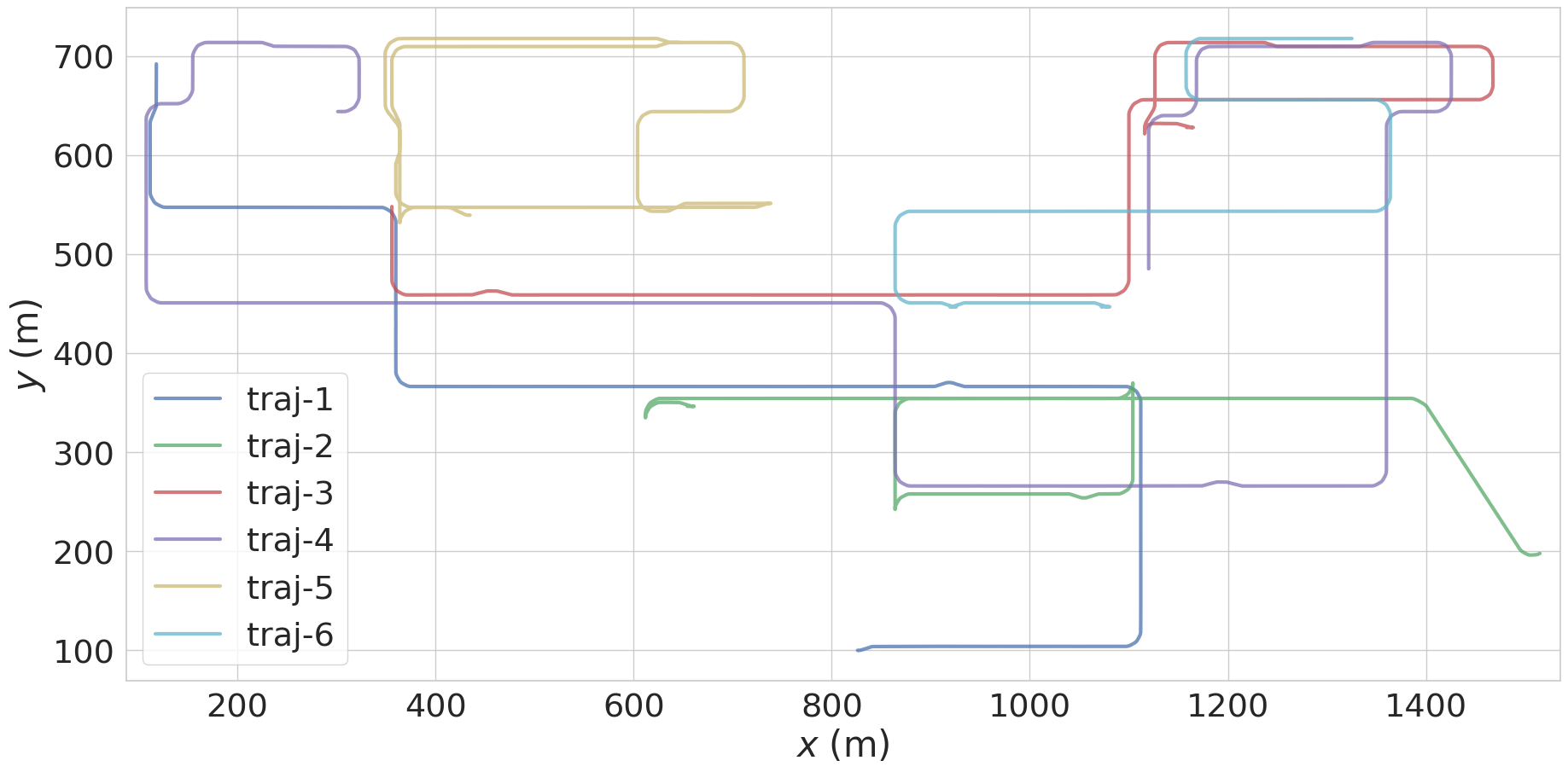}}
    \hfill
  \subfloat[\label{fig:trajectories}]{%
       \includegraphics[width=\linewidth]{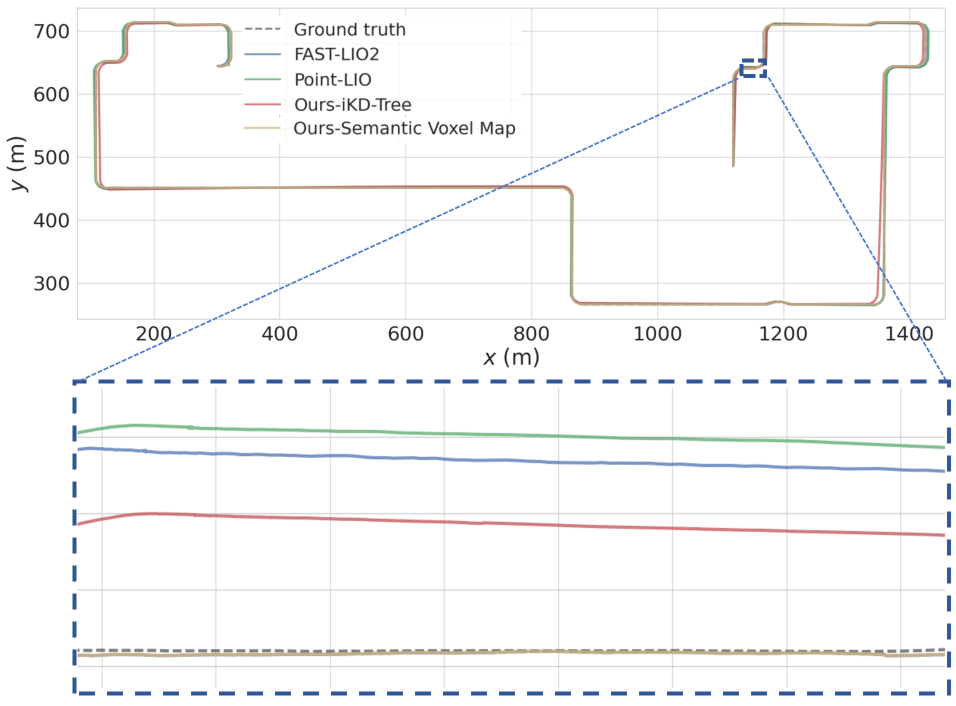}}
    \hfill
    % 总标题
    \caption{(a) The trajectories on our private dataset are demonstrated. These 6 trajectories cover multiple distinct motion scenarios, primarily traversing the main passageways of an industrial port. (b) Detailed comparison of the longest trajectory (\textbf{traj-4}) from (a), illustrating the performance of FAST-LIO2, Point-LIO, our primary proposed method (Ours-Semantic Voxel Map), and its variant (Ours-iKD-Tree). This comparison emphasizes long-term accuracy and robustness across varying levels of environmental complexity.}
    % \label{fig:maps}
    \vspace{-1em}
\end{figure}
%%%%%%%%%%%%%%%%%%%%%%%%%%%%%%%

\subsection{Experiment Setup}
\label{sec:exp_set}

\noindent\textbf{Global Prior Map Construction}. 
Port scenario is large-scale, highly dynamic, and unbounded, with sparse yet repetitive environment features. Frequent movements of mechanical equipment, dense vehicle occlusions, and complex logistics operations result in significant temporal variations in environmental features. Besides, LiDAR data primarily captures ground information, posing challenges for stable map construction and high-precision localization.
In global map construction, we focus on extracting static features (e.g., ground elements) and systematically eliminate potential dynamic objects  from the map.
Additionally, the structured static semantic features (such as pillars, as shown in Fig.~\ref{fig:pillar}) present in the environment are incorporated to enhance environmental understanding and ensure accurate and stable global pose constraints. Finally, a stable and reliable global prior map covering approximately 1 million square meters is generated, as illustrated in Fig.~\ref{fig:maps}.
%, enabling robust and precise localization in dynamic and complex port scenarios.
% To address the challenges in port scenarios and improve scan-to-map matching efficiency, extensive validation and optimization in real-world scenarios are conducted. Ultimately, we prioritized the extraction of static features (such as ground elements) and systematically removed all potential dynamic objects from the map. 
% Then, the environment is voxelized in 0.5-meter grid size to refine ground feature representation, using the distribution of point clouds within each voxel. 
% To better extract fine and subtle ground features, the entire environment is voxelized, with the distribution of point clouds within each voxel used to represent environmental features. 

\begin{table*}[]
\caption{
    % \MakeUppercase{Quantitative comparison across six datasets (corresponding to each trajectory in Fig. \ref{fig:trajectories}). }
    Quantitative comparison for Light-LOAM \cite{10439642}, FAST-LIO2 \cite{9697912}, Point-LIO \cite{Point-LIO}, our primary proposed method (Ours-Semantic Voxel Map), and its variant (Ours-iKD-Tree) across six datasets (corresponding to each trajectory in Fig. \ref{fig:top}). The dash $``-"$ indicates that the method \textit{\textbf{fails}} on this test trajectory.
}
\label{tab:quantitative-results}
\resizebox{\textwidth}{!}{%
\begin{tabular}{@{}ccccccccc@{}}
\toprule
Dataset &
  Mileage (km) &
  Case &
  % Max absolute pose error (m) &
  % Mean absolute pose error (m) &
  % Max lateral error (m) &
  % Mean lateral error (m) &
  % Max longitudinal error (m) &
  % Mean longitudinal error (m) \\ 
  \makecell{Max absolute \\ \;pose error (m)} &
  \makecell{Mean absolute \\ pose error (m)} &
  \makecell{Max lateral \\ error (m)} &
  \makecell{Mean lateral \\ error (m)} &
  \makecell{Max longitudinal \\ error (m)} &
  \makecell{Mean longitudinal \\ error (m)} \\
  \midrule
\multirow{6}{*}{\textbf{traj-1}} & \multirow{6}{*}{1.843} & Light-LOAM  & $-$ & $-$ & $-$ & $-$  & $-$ & $-$ \\  % & \hl{444.196-} & 427.132 & 426.447 & 22.637  & 438.021 & 0.234 \\
                            &                        & FAST-LIO2               & 4.255 & 2.379 & 3.893 & 1.189  & 4.097 & \textbf{0.001} \\
                            &                        & Point-LIO               & 21.283 & 1.676 & 2.585 & 0.829  & 21.257 & 0.013 \\
                            &                        & Ours-iKD-Tree           & 10.932 & 5.583 & 9.721 & 2.729  & 10.797 & \textbf{0.001} \\
                            % &                        & Ours-Voxel Map          & \textbf{0.274} & \textbf{0.129} & \textbf{0.196} & 0.040  & \textbf{0.257} & 0.064 \\
                            &                        & Ours-Semantic Voxel Map & \textbf{0.278} & \textbf{0.129} & \textbf{0.198} & \textbf{0.039}  & \textbf{0.265} & 0.064 \\ \midrule
\multirow{6}{*}{\textbf{traj-2}} & \multirow{6}{*}{1.770} & Light-LOAM  & $-$ & $-$ & $-$ & $-$  & $-$ & $-$ \\  % & 442.882 & 421.955 & 335.217 & 256.280  & 442.833 & 301.822 \\
                            &                        & FAST-LIO2               & 3.421 & 0.638 & 3.239 & 0.283  & 2.647 & 0.013 \\
                            &                        & Point-LIO               & 10.681 & 0.377 & 2.266 & 0.119  & 10.450 & 0.094 \\
                            &                        & Ours-iKD-Tree           & 0.476 & 0.156 & 0.440 & 0.043  & \textbf{0.370} & \textbf{0.001} \\
                            % &                        & Ours-Voxel Map          & \textbf{0.374} & \textbf{0.120} & \textbf{0.346} & 0.028  & 0.372 & 0.008 \\
                            &                        & Ours-Semantic Voxel Map & \textbf{0.374} & \textbf{0.121} & \textbf{0.346} & \textbf{0.027}  & 0.372 & 0.009 \\ \midrule
\multirow{6}{*}{\textbf{traj-3}} & \multirow{6}{*}{1.914} & Light-LOAM         & \textbf{0.621} & 0.400 & 0.319 & 0.090  & \textbf{0.598} & \textbf{0.001} \\
                            &                        & FAST-LIO2               & 5.053 & 2.554 & 4.595 & 1.194  & 4.965 & 1.240 \\
                            &                        & Point-LIO               & 6.606 & 2.029 & 3.823 & 0.958  & 6.394 & 0.866 \\
                            &                        & Ours-iKD-Tree           & 13.486 & 5.314 & 12.882 & 2.779  & 13.196 & 2.331 \\
                            % &                        & Ours-Voxel Map          & 0.711 & \textbf{0.128} & \textbf{0.272} &  \textbf{0.034} & 0.705 & 0.030 \\
                            &                        & Ours-Semantic Voxel Map & 0.699 & \textbf{0.132} & \textbf{0.302} & \textbf{0.041}  & 0.693 & 0.032 \\ \midrule
\multirow{6}{*}{\textbf{traj-4}} & \multirow{6}{*}{2.950} & Light-LOAM  & $-$ & $-$ & $-$ & $-$  & $-$ & $-$ \\  %& 187.641 & 87.617 & 180.913 &  12.143 & 158.569 & 50.894 \\
                            &                        & FAST-LIO2               & 4.869 & 2.654 & 4.346 & 0.562  & 4.864 & 1.439 \\
                            &                        & Point-LIO               & 14.312 & 3.242 & 5.065 & 0.741  & 14.276 & 1.778 \\
                            &                        & Ours-iKD-Tree           & 11.061 & 3.572 & 10.979 & 0.139  & 10.825 & 1.868 \\
                            % &                        & Ours-Voxel Map          & \textbf{0.518} & \textbf{0.162} & \textbf{0.307} & \textbf{0.043}  & \textbf{0.518} & 0.094 \\
                            &                        & Ours-Semantic Voxel Map & \textbf{0.669} & \textbf{0.164} & \textbf{0.309} & \textbf{0.044}  & \textbf{0.669} & \textbf{0.092} \\ \midrule
\multirow{6}{*}{\textbf{traj-5}} & \multirow{6}{*}{1.840} & Light-LOAM & $-$ & $-$ & $-$ & $-$  & $-$ & $-$ \\  % & 152.510 & 72.132 & 148.813 & 1.040  & 148.185 & 13.243 \\
                            &                        & FAST-LIO2               & 1.202 & 0.378 & 0.721 & 0.087  & 1.133 & 0.245 \\
                            &                        & Point-LIO               & 3.619 & 0.825 & 1.231 & 0.044  & 3.619 & 0.399 \\
                            &                        & Ours-iKD-Tree           & 1.518 & 0.613 & 1.246 & 0.162  & 1.487 & 0.348 \\
                            % &                        & Ours-Voxel Map          & 0.983 & \textbf{0.128} & 0.203 & \textbf{0.016}  & 0.974 & \textbf{0.086} \\
                            &                        & Ours-Semantic Voxel Map & \textbf{0.483} & \textbf{0.129} & \textbf{0.201} & \textbf{0.018} &  \textbf{0.479} & \textbf{0.090} \\ \midrule
\multirow{6}{*}{\textbf{traj-6}} & \multirow{6}{*}{1.368} & Light-LOAM & $-$ & $-$ & $-$ & $-$  & $-$ & $-$ \\  % & 182.346 & 102.505 & 120.061 & 58.409  & 137.242 & 65.996 \\
                            &                        & FAST-LIO2               & 1.044 & 0.431 & 1.042 & 0.098  & 0.999 & 0.047 \\
                            &                        & Point-LIO               & 18.224 & 0.535 & 1.092 & 0.101  & 18.214 & 0.113 \\
                            &                        & Ours-iKD-Tree           & \textbf{0.397} & 0.118 & 0.252 & 0.039  & \textbf{0.322} & 0.058 \\
                            % &                        & Ours-Voxel Map          & \textbf{0.375} & \textbf{0.076} & 0.157 & \textbf{0.027}  & 0.374 & 0.043 \\
                            &                        & Ours-Semantic Voxel Map & 0.422 & \textbf{0.076} & \textbf{0.142} & \textbf{0.027}  & 0.420 & \textbf{0.042} \\ \bottomrule
\end{tabular}%
}
% \vspace{4.5 em}
% \footnotesize % 设置较小的字体
% The dash ``--'' indicates that the method \textit{\textbf{fails}} on this test trajectory.
\vspace{-2em}
\end{table*}

\noindent\textbf{Datasets and Sensor Equipments}.
% To evaluate the performance of our method in large-scale dynamic environments, we selected a challenging port scenario for testing. 
% % Port environments present numerous challenges, including frequent machinery movements, dense vehicle occlusions, sparse and repetitive environmental features, and GPS signal interference caused by metal structures. 
% Port environments are highly dynamic, large-scale, and open, with sparse yet repetitive environment features. Frequent movements of mechanical equipment, dense vehicle occlusions, and complex logistics operations result in significant temporal variations in environmental features. Besides, LiDAR data primarily captures ground information, posing challenges for stable map construction and high-precision localization. 
As illustrated in Fig. \ref{fig:igv}, each IGV is equipped with a multi-sensor layout that includes two diagonally mounted 16-line LiDARs (10\,Hz) to maximize coverage for environmental perception, a centrally located INS/IMU (20\,Hz) for attitude and motion information, two GNSS antennas mounted at the front-left and rear-left corners for high-precision positioning, and four wheel encoders (50\,Hz) and four steering angle sensors (50\,Hz) located at the center of each axle for vehicle motion estimation and odometry calculation. 
The LiDAR point clouds are synchronized, distortion-corrected, and then transformed into the vehicle’s central coordinate system. 
Wheel odometry data (50\,Hz) is generated by fusing inputs from wheel encoders, steering angle sensors, and motor feedback signals using a vehicle kinematic model. This data is further refined through corrections from IMU pre-integration measurements and transformed into the vehicle’s central coordinate system. 
High-precision localization results, based on an RTK-aided multi-sensor fusion approach (50\,Hz), are used as ground truth for quantitative performance evaluation, ensuring the reliability and accuracy of the system’s localization and navigation.
We manually construct a comprehensive dataset covering approximately 1 square kilometer, consisting of six test trajectories collected from six IGVs, with a total length of 11.685 kilometers (trajectories are showed in Fig.~\ref{fig:top}, while detailed mileage data is provided in the Table \ref{tab:quantitative-results}).

\noindent\textbf{Implementation Platform and Hardware}. 
All our comparative experiments are conducted on a computer equipped with an Intel i7-8750H CPU (2.20 G\,Hz), 32 GB of RAM, and a GeForce GTX 1050 Ti GPU, running the Ubuntu 20.04 operating system. The experimental code is efficiently written in C++ and developed based on the Robot Operating System (ROS) to ensure modularity and flexibility, facilitating testing and system integration.

\noindent\textbf{Baselines and Metrics.} 
To validate the superiority of our proposed method, we conduct comparative experiments to comprehensively evaluate its advantages in terms of localization reliability. We have select several state-of-the-art and representative baseline methods, including FAST-LIO2 \cite{9697912}, Point-LIO \cite{Point-LIO}, and Light-LOAM \cite{10439642}. To ensure fairness in comparison, the experimental results of all baseline methods are obtained using the original source code provided by their authors, with only minor adjustments to the input data interface to adapt to our dataset. 
Additionally, to further demonstrate the advantages of our proposed map representation, we compare the conventional iKD-Tree-based map with our proposed formats: the voxelized map enriched with semantic information, referred to as Ours-iKD-Tree and Ours-Semantic Voxel Map, respectively, as shown in the Table \ref{tab:quantitative-results}. 
Absolute Trajectory Error (ATE) is adopted as the primary metric for evaluating SLAM accuracy, and the $evo$ toolkit is utilized to analyze and compare the localization trajectories of different algorithms.

\subsection{Localization Results and Comparison}
\label{sec:loc_result}

The full test trajectories are illustrated in Fig.~\ref{fig:top}. Light-LOAM \cite{10439642} is found to fail on most port test datasets, displaying significant errors. Therefore, the analysis primarily focuses on the successful algorithms: Point-LIO \cite{Point-LIO}, FAST-LIO2 \cite{9697912}, and our primary proposed method (Ours-Semantic Voxel Map), and its variant (Ours-iKD-Tree). 

\noindent\textbf{Trajectory Comparison.} 
Our proposed methods demonstrate close alignment with ground truth trajectories across all 6 test datasets. 
% Among these, Ours-Semantic Voxel Map and Ours-Voxel Map achieve the highest accuracy and robustness.
This performance is exemplified by the longest trajectory traj-4 (over 2.950 km) in Fig. \ref{fig:trajectories}. In this large-scale scenario,
FAST-LIO2 \cite{9697912} and Point-LIO \cite{Point-LIO} accumulate significant drift and exhibit instability over the long distance.
Ours-iKD-Tree improves accuracy but lacks long-term stability compared to the semantic voxel-based approach.
Ours-Semantic Voxel Map show exceptional performance, with enhanced stability, particularly in dynamic and cluttered environments, demonstrating the critical role of semantic information for robust, long-term localization.
%Across six test datasets, our proposed methods closely align with the ground truth trajectories, with Ours-Semantic Voxel Map and Ours-Voxel Map achieving the highest accuracy and robustness across all scenarios. 
%Taking \textbf{traj-4}, the longest trajectory spanning 2.950 km, from Fig. \ref{fig:trajectories} as an example, illustrated in Fig.~\ref{fig:traj-4},  FAST-LIO2 \cite{9697912} and Point-LIO \cite{Point-LIO} suffer from significant drift and instability over long distances in this complex scenario. Ours-iKD-Tree improves accuracy but lacks long-term stability compared to voxel-based methods. Both Ours-Voxel Map and Ours-Semantic Voxel Map demonstrate exceptional performance, with the latter further enhancing stability in dynamic and cluttered environments, highlighting the critical role of semantic information in achieving robust and consistent long-term localization.

\noindent\textbf{Pose Error Quantitative Comparison.} 
Table \ref{tab:quantitative-results} highlights notable differences in localization errors across six algorithms. Light-LOAM exhibits the highest errors and poorest adaptability across various scenarios, underscoring its lack of robustness. 

In simple scenarios, FAST-LIO2 achieves a maximum error of 1.044\,m and an average of 0.431\,m (\textbf{traj-6}), while Point-LIO shows a maximum error of 10.681\,m and an average of 0.377\,m (\textbf{traj-2}). In complex environments, FAST-LIO2's maximum error rises to 5.053\,m (\textbf{traj-3}), and Point-LIO's increases to 14.312\,m (\textbf{traj-4}), both showing significant robustness degradation.

Ours-iKD-Tree demonstrates significantly better performance compared to Light-LOAM, FAST-LIO2 and Point-LIO. For example, in \textbf{traj-2}, its maximum absolute error is 0.476\,m, with an average error of 0.156\,m, a substantial improvement over FAST-LIO2 (3.421\,m), and Point-LIO (10.681\,m). However, in certain scenarios, such as \textbf{traj-3} and \textbf{traj-4}, its maximum errors are relatively high (13.486\,m and 11.061\,m, respectively), indicating room for improvement in dynamic or long-distance tasks. 

Ours-Semantic Voxel Map consistently outperforms Ours-iKD-Tree across all scenarios, demonstrating superior robustness. For instance, in \textbf{traj-1}, its maximum error is only 0.278\,m, with an average error of 0.129\,m, significantly better than Ours-iKD-Tree (10.932\,m and 5.583\,m). In \textbf{traj-4}, its maximum error is 0.669\,m, with an average error of 0.164\,m, compared to Ours-iKD-Tree (11.061\,m and 3.572\,m), showing remarkable improvement.

% Ours-Voxel Map consistently outperforms Ours-iKD-Tree across all scenarios, demonstrating superior robustness. For instance, in \textbf{traj-1}, its maximum error is only 0.274\,m, with an average error of 0.129\,m, significantly better than Ours-iKD-Tree (10.932\,m and 5.583\,m). In \textbf{traj-4}, its maximum error is 0.518\,m, with an average error of 0.162\,m, compared to Ours-iKD-Tree's 11.061\,m and 3.572\,m, showing remarkable improvement.

% With semantic information integrated, Ours-Semantic Voxel Map performs similarly to Ours-Voxel Map in simple scenarios but exhibits clear advantages in dynamic and complex environments. For example, in \textbf{traj-3}, its maximum error decreases from 0.711\,m to 0.699\,m, and in \textbf{traj-5}, the maximum error drops significantly from 0.983\,m to 0.483\,m, demonstrating enhanced robustness. Therefore, Ours-Semantic Voxel Map is the ideal solution for dynamic, complex environments and long-distance tasks.


\section{CONCLUSION}

Building on the iESKF filter, this paper proposes a tightly-coupled semantic LiDAR-inertial-wheel odometry fusion framework. Within this framework, a semantic voxel-based matching algorithm is introduced to effectively distinguish and leverage diverse semantic features, thereby mitigating long-term trajectory drift. 
Additionally, a 3D adaptive scaling strategy is presented to optimize wheel odometry performance on complex terrains. Extensive experiments demonstrate the superiority of our method over state-of-the-art approaches in dynamic and large-scale environments. 
Successfully deployed in a one-million-square-meter automated port, the system delivers precise and stable localization for 35 IGVs, demonstrating reliability in real-world applications.
Future work will focus on expanding semantic element detection, supporting objects with distinctive geometric shapes, and incorporating visual semantic information to further enrich the semantic content of the map and enhance the algorithm's performance in texture-sparse scenarios.

\bibliographystyle{IEEEtran}
\bibliography{IEEEabrv,IEEEexample,mybibfile}

\end{document}